\documentclass[conference]{IEEEtran}
\IEEEoverridecommandlockouts
\usepackage{cite}
\usepackage{amsmath,amssymb,amsfonts}
\usepackage{algorithmic}
\usepackage{graphicx}
\usepackage{stfloats} 
\usepackage{tcolorbox}
\usepackage{textcomp}
\usepackage{xcolor}
\usepackage{tabularx}
\usepackage{hyperref}
\def\BibTeX{{\rm B\kern-.05em{\sc i\kern-.025em b}\kern-.08em
    T\kern-.1667em\lower.7ex\hbox{E}\kern-.125emX}}

\title{Australian Bushfire Intelligence with AI-Driven Environmental Analytics
}

\author{
Tanvi Jois$^{\ast}$, Hussain Ahmad$^{\ast}$, Fatima Noor$^{\dagger}$, Faheem Ullah$^{\ddagger}$%
\thanks{Author Emails: 
tanvi.jois@student.adelaide.edu.au (Tanvi Jois); 
hussain.ahmad@adelaide.edu.au (Hussain Ahmad); 
fatimanoorr39@gmail.com (Fatima Noor); 
faheem.ullah@zu.ac.ae (Faheem Ullah)}%
\\[0.3em]
$^{\ast}$Adelaide University, Australia\\
$^{\dagger}$University of Engineering and Technology, Pakistan\\
$^{\ddagger}$Zayed University, UAE
}
\begin{document}
\maketitle

\thispagestyle{plain}
\pagestyle{plain}

\begin{abstract}

Bushfires are among the most destructive natural hazards in Australia, causing significant ecological, economic, and social damage. Accurate prediction of bushfire intensity is therefore essential for effective disaster preparedness and response. This study examines the predictive capability of spatio-temporal environmental data for identifying high-risk bushfire zones across Australia. We integrated historical fire events from NASA FIRMS, daily meteorological observations from Meteostat, and vegetation indices such as the Normalized Difference Vegetation Index (NDVI) from Google Earth Engine for the period 2015–2023. After harmonizing the datasets using spatial and temporal joins, we evaluated several machine learning models, including Random Forest, XGBoost, LightGBM, a Multi-Layer Perceptron (MLP), and an ensemble classifier. Under a binary classification framework distinguishing “low” and “high” fire risk, the ensemble approach achieved an accuracy of 87\%. The results demonstrate that combining multi-source environmental features with advanced machine learning techniques can produce reliable bushfire intensity predictions, supporting more informed and timely disaster management.

\end{abstract}


\section{Introduction}
Bushfires rank among the most severe natural disasters in Australia, causing widespread damage \cite{fb1456}. Predictive modeling of bushfires can provide disaster management services with the necessary tools to improve allocation of resources and reduce the potential catastrophic impacts of the bushfires. Recent accounts of the 2019-2020 "Black Summer fires" indicate that close to 19 million hectares were burnt, and there was strong evidence that anthropogenic climate change contributed to the increase in the chance of fire weather conditions in south-eastern Australia \cite{vanb29}. This has solidified the perception of bushfires as an escalating ecological hazard with growing ecological, economic, and public health impacts. Rapid progress in artificial intelligence \cite{abbas2025, abbas2024robust}, particularly large language models (LLMs) \cite{chopra2024chatnvd, ahmad2025future}, has transformed fields such as cybersecurity \cite{abdulsatar2025towards, goel2024machine, ahmad2025survey, goel2025co, jayalath2024microservice, ullah2025skills}, finance systems \cite{zhang2025regimefolio, chen20253s}, and cloud computing \cite{ahmad2024smart, ahmad2025towards, ahmad2025resilient}. Motivated by these advances, recent studies have increasingly leveraged machine learning and deep learning for wildfire prediction and fire spread modelling, demonstrating both their potential and their current limitations \cite{b30}. Furthermore, work on wildfire susceptibility mapping using multi-sensor satellite data combined with machine-learning and deep-learning models also shows the use of spatially explicit and data-driven methods as the foundation to operational decision-making and landscape-scale bushfire management \cite{b31}. Recent Australian studies emphasize the importance of evidence-based decision-making in bushfire management, supporting the need for data-driven risk and intensity modeling\cite{b33}. Beyond fire occurrence, recent work also shows how fire interacts with subsequent rainfall and runoff to shape downstream environmental impacts, indicating why intensity-aware modeling is valuable for proactive environmental management and response planning \cite{b34}.

Despite various efforts to predict bushfires using meteorological data and remote sensing data, it remains a complex problem to solve. Bushfire occurs due to various climatic, geographic, and vegetation factors that vary significantly across time and space. Natural data are always heavily imbalanced as fire occurrences are rare compared to non-fire events. Even the relationships between environmental variables are non-linear, adding further complexity. To address such challenges, this project aims to investigate spatio-temporal climatic data, to train machine learning models to identify zones of high bushfire intensity in Australia.
Recent studies have explored machine learning approaches for wildfire prediction, but several limitations continue to restrict the usefulness of such models \cite{b23,b24,b25,b28}.

Existing wildfire prediction and vulnerability studies provide important foundations, but still leave several gaps that are directly relevant to this work. 
One study predicts year-round daily wildfire occurrence in Gangwon state South Korea using topographic factors without specifically taking into account fire intensity metrics like Fire Radiative Power \cite{b23}. Other works generate yearly maps of wildfire vulnerability using random forest and logistic regression; however, they do not capture fine-grained spatiotemporal dynamics or estimate intensity-based fire risk \cite{b24}. Ensemble-based approaches have also been applied to estimate forest-fire risk in Yunnan Province, China; however, these methods rely on static datasets and do not combine vegetation indices, climate factors, and multi-year fire events into a single pipeline \cite{b25}. Similarly, deep learning models using satellite imagery have been used to map wildfire vulnerability \cite{b28}, yet they still lack multi-source, multi-year event-level fusion of vegetation time series, meteorological information, and FIRMS. These limitations motivate the need for an integrated, multi-source, and multi-year framework that combines vegetation indices, meteorological data, and intensity-based fire event features to model wildfire risk more comprehensively. 

\begin{table*}[!b]
\centering
\caption{Comparison of Key Features in Bushfire-Related Studies and the Present Work}
\label{tab:lit_comparison}
\begin{tabular*}{\textwidth}{@{\extracolsep{\fill}} l c c c c c c c c}
\hline
\textbf{Study} &
\textbf{Fire events} &
\textbf{Veg.\ indices} &
\textbf{Climate / met.} &
\textbf{Remote sensing} &
\textbf{Multiple ML} &
\textbf{Ensemble} &
\textbf{Accuracy} &
\textbf{Multi-year}  \\
\hline
Partheepan et al.~\cite{b2} & \checkmark & \checkmark & \checkmark & \checkmark & -- & -- & 86.13\% & \checkmark \\
Michael et al.~\cite{b3}     & -- & \checkmark & \checkmark & \checkmark & --         & -- & 86.13\%  & -- \\
Demir~\cite{b4}              & -- & \checkmark & --         & \checkmark & --         & -- & --       & -- \\
Hosseini \& Lim~\cite{b5}    & \checkmark & \checkmark & \checkmark & \checkmark & \checkmark & -- & 82--85\% & -- \\
Bandara et al.~\cite{b6}     & -- & \checkmark & \checkmark & \checkmark & \checkmark & -- & $\sim$85\% & -- \\
\textbf{Our work}       & \checkmark & \checkmark & \checkmark & \checkmark & \checkmark & \checkmark & 87\% & \checkmark \\
\hline
\end{tabular*}

\end{table*}

This study aims to answer the question “Can spatio-temporal, historical, and time-series data be used to develop a model that predicts zones of high bushfire intensity across different environmental conditions in Australia?” To address this question, the project integrated three different datasets. Historical fire events were obtained from NASA’s Fire Information for Resource Management System (FIRMS) \cite{b7}. Daily weather data were collected from Meteostat \cite{b8}, and Google Earth Engine’s (GEE) \cite{b985} Normalized Difference Vegetation Index (NDVI) data were collected for vegetation information during the period 2015-2023. Several machine learning models, including Light GBM, XGBoost, Random Forest, and a Multi-layered Perceptron (MLP), as well as an ensemble model, were tested for two-class classification of the fire risk. Models were evaluated on various metrics.

In summary, our paper makes the following contributions:
\begin{itemize}
\item Design and implementation of a spatio-temporal modeling system integrating bushfire event data, weather data, and vegetation indices.
\item Implementation of advanced feature engineering, class imbalance handling, and ensemble modeling to capture the complexity of the bushfire data.
\item Evaluation and comparison of multiple machine learning models to identify the best performing method for disaster management systems.
\end{itemize}

The paper is structured as follows: Section \ref{sec:litreview} summarizes the related work and provides a comparison of the proposed work with the existing literature. Section \ref{sec:method} describes our research methodology, including data collection, preprocessing, feature engineering, and class construction. Section \ref{sec:experiments} reports the experimental evaluation of this study. Section \ref{sec:discussion} interprets the results, while Section 
\ref{sec:limitations} and  Section \ref{sec:conclusion} outline the threats to validity and conclusion, respectively.

\section{Related Work} \label{sec:litreview}
This section reviews prior research on bushfire prediction using meteorological variables, vegetation indices, remote sensing data, and machine learning techniques. It highlights key methodological advances and limitations in prior studies, establishing the research gap that motivates the present work.

Bushfire prediction is active, with approaches leveraging remote sensing, meteorological data, machine learning, and citizen science. Prior studies demonstrate progress but also highlight the challenges of handling imbalanced datasets and data integration. Recent work also shows that bushfires can drive substantial environmental impacts after the fire, such as degraded water quality due to runoff after rain. This justifies the significance of effective bushfire risk and intensity modeling in environmental management, rather than focusing only on post-fire response\cite{b35}.

Table \ref{tab:lit_comparison} compares the main features used in previous bushfire studies with those in the present work. Most earlier studies combined vegetation indices, climate data, and remote sensing; however fewer studies integrated multi-year event dataset or ensemble models. 

\begin{table*}[!t]
\centering
\caption{Comparison of Studies: Data Sources and Vegetation Indices}
\label{tab:vegetation_models}

\begin{tabular}{|l|p{5cm}|p{5cm}|}
\hline
\textbf{Study} & \textbf{Data Sources} & \textbf{Vegetation Indices} \\
\hline
Partheepan et al.~\cite{b2} & FIRMS + Landsat (GEE) + ERA5-Land + SMAP + SRTM & NDVI, NBR, dNBR, EVI, NDWI, BI \\
\hline
Michael et al. \cite{b3} & Landsat + Climatic + Topographical & NDVI, NBR \\
\hline
Demir (2020) \cite{b4} & Landsat 8/5 + GEE NDVI & NDVI \\
\hline
Hosseini \& Lim \cite{b5} & MODIS + Climatic + Topographic + Anthropogenic & NDVI, SAVI, NBR \\
\hline
Bandara et al. \cite{b6} & Mediterranean climate + Topography + Vegetation & NDVI, NDMI \\
\hline
\textbf{Our Work} & FIRMS + Meteostat + GEE NDVI & NDVI only \\
\hline
\end{tabular}%

\end{table*}

Pratheepan et al.\cite{b2} investigated bushfire severity modelling and the fire trends across Australia by integrating multi-source remote sensing data with machine learning. The study utilised NASA FIRMS for a 12 year period from 2012-2023 alongside the Landsat 5,7,8 images processed via Google Earth Engine to extract vegetation and burn-related information. The key indices included Enhanced Vegetation Index(EVI), Normalized Difference Vegetation Index (NDVI), Normalized Burn Ratio (NBR), difference Normalized Burn Ratio (dNBR), EVI, Normalized Difference Water Index (NDWI) and Burn Index (BI) which were combined with topographical variables like Soil Moisture Active Passive (SMAP) and Shuttle Radar Topography Mission (SRTM). Climatic factors including temperature and precipitation from ERA5-Land were also used. An XGBoost regression model was developed using dNBR as the target variable to quantify fire severity achieving an R$^2$ of 0.86. The results showed the dominant role of burn sensitive bands and enabled identification of high-risk fire zones in Australia. 

Michael et al.\cite{b3} explored the integration of remote sensing data and meteorological data for the prediction of bushfire risk. 
The authors used the combination of Landsat-derived indices such as NDVI and NBR with topographical and climatic variables. Models like XGBoost were employed to achieve an accuracy of 86.13\%. They discussed how predicted output could support resource allocation analysis included the fire risk prediction with data-driven damage management resource allocation for fire fighting and damage mitigation. They also discussed the integration of real-time data sourced from a high-powered unmanned aerial vehicle (UAV) as a future improvement for operational bushfire predictions.

The Australian Bushfires of 2019-2020 was analysed by another study using NDVI and GEE data for rapid vegetation cover assessment \cite{b4}. Landsat 8 and 5 data were processed to monitor vegetation changes with a threshold value of 0.1509 for the NDVI to estimate the forest cover. The results showed a significant loss of vegetation. However, even with such destruction, vegetation recovery was observed between January and February 2020. The study highlighted the effectiveness of a cloud-based platform and real-time fire damage monitoring system and concluded that remote sensing is crucial for environmental bushfire assessment.

In Southern Australia, Hosseini and Lim \cite{b5} investigated bushfire risk using multi-source data and several machine learning models. The authors combined vegetation indicators such as NDVI with climatic, topographic, and human-related factors. Multiple classifiers were tested, including Random Forest, Support Vector Machine, Frequency Ratio, and logistic regression. Random Forest achieved the highest accuracy of about 85\%, while SVM reached around 82\%. 

Wildfire prediction using climate, vegetation and topographical variables in a Mediterranean environment was developed by Bandara et al. \cite{b6}. The authors combined indices such as NDVI and Normalised Difference Moisture Index (NDMI), topographic data with meteorological predictors to assess fire risk. Multiple models like logistic regression, random forest and support vector machines were used and evaluated. Random forest achieved the highest performance with an accuracy close to 85\% effectively identifying fire risk zones. The importance of vegetation moisture was shown in this study while predicting fire risks.

\subsubsection*{Distinguishing Features of the Present Work}
In contrast to previous studies, this project involves a streamlined approach by focusing on the integration of NASA’s FIRMS dataset with Meteostat daily weather variables and GEE’s NDVI data between 2015-2023 \cite{b7}\cite{b8}. Unlike several studies that employed a broad range of vegetation indices, this project is concentrated only on NDVI for better computational feasibility. Furthermore, the present study explored two-class classification (low and high). The two-class approach achieved an accuracy of 87\%. Another major factor is the modeling approach. Unlike most studies, this project compared multiple models like random forest, MLP, XGBoost and LightGBM and then created an ensemble model to achieve the best results, guaranteeing robustness and adaptability.

To give a comparison of previous work, Table~\ref{tab:vegetation_models} demonstrates how various combinations of remote sensing, vegetation indices, meteorological variables, and topographic data were used in earlier studies, frequently without multi-year event data or intensity modeling.  By contrast, our work incorporates vegetation, climate, and fire history into a single spatiotemporal pipeline.  By focusing on predicting the possible intensity of fire rather than just its presence, the application of FRP further sets the model apart. Although previous research employed just one or two models, our work assesses a wide range of machine learning techniques, such as Random Forest, MLP, XGBoost, and LightGBM, followed by a stacked ensemble to obtain robust performance. The final two-class ensemble model addresses the imbalance issue that many earlier research ignore, achieving 87\% accuracy and showing enhanced sensitivity toward high-risk fires.  The suggested method is more practically dependable and applicable than previous studies due to the complete incorporation of data sources, intensity-based categorization, and ensemble learning.

\begin{figure*}[t]
    \centering
    \includegraphics[height=0.3\textheight, keepaspectratio]{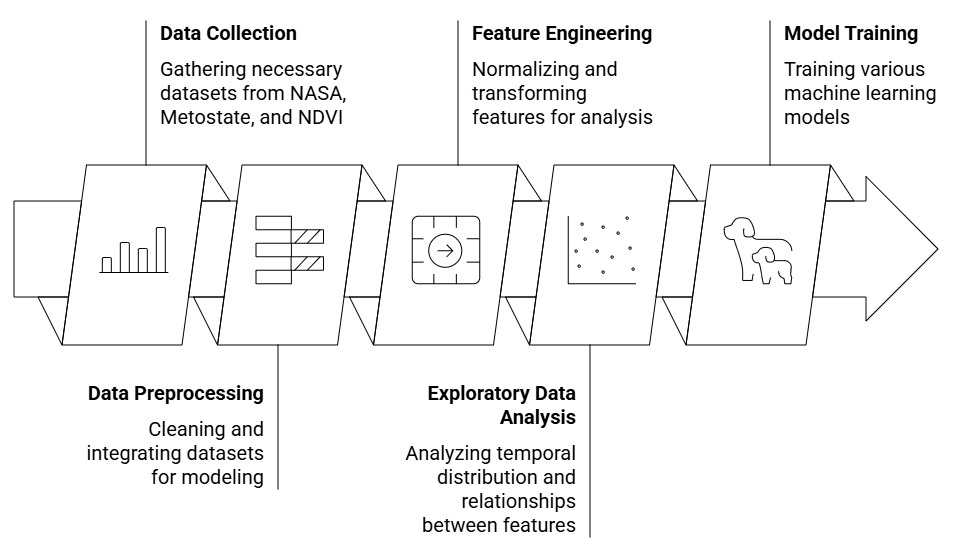}
    \caption{Research Methodology.}
    \label{fig:workflow}
\end{figure*}

\section{Research Methodology} \label{sec:method}
This section reports the research methodology adopted in this paper. The study is divided into five systematic phases. The workflow was designed to ensure methodological and practical relevance. As illustrated in Fig~\ref{fig:workflow}, the workflow consists of five phases beginning with data collection and ending with model training.

\subsection{Data Collection}\label{sec:data_collection}
The first step involved collecting a dataset capturing fire events, climate conditions, and vegetation dynamics for the 2015-2023 period.
\begin{enumerate}
    \item NASA FIRMS: The NASA FIRMS data was obtained from NASA Earthdata platform \cite{b7} offering a reliable geolocated fire event data, supplying historical record of bushfire activity.

    \item Meteostat: Daily weather observations across major stations in Australia was collected through
    Meteostat platform \cite{b8} providing valuable climatic information.
    
    \item NDVI: The GEE was used to collect NDVI data using Moderate Resolution Imaging Spectroradiometer (MODIS) satellite products. NDVI provides effective values for vegetation health, which is a crucial factor in bushfire ignition and spread \cite{b9}.
\end{enumerate}

The combination of these datasets with relevant features allowed the perspective of a spatio-temporal context within environmental and climatic conditions.

\subsection{Data Preprocessing}
The raw datasets obtained from the FIRMS, Meteostat and GEE were cleaned extensively to obtain datasets that were ready for modeling.
\paragraph{Spatial and Temporal join}

Each fire event in FIRMS has latitude, longitude and timestamp values for the period 2015-2023. Meteostat however is daily major station data. This does not align with FIRMS fire detections. To address this issue, a 5 km search radius was used to assign each fire event to the nearest available weather station. Multiple weather stations in the radius could cause complexity in the model and hence, an inverse-distance weighted (IDW) interpolation was inculcated to reduce single station bias \cite{b10}.
The latitudes and longitudes in FIRMS data were filtered to cover only Australia. Using the latitude and longitude values, the coordinates were geographically classified into regions matching the Meteostat major station (Australian) weather data. The integration was performed based on the region and timestamp.
NDVI values (approx. 250 m resolution MODIS) were sampled directly from GEE to match the fire event coordinates ensuring the vegetation conditions correspond to the exact fire location. Since NDVI data was obtained in a 16-day interval, an ±8 day window was accounted to align each fire event with the closest available NDVI composite. If  NDVI was unavailable, the record was excluded. 

\paragraph{Data Cleaning}
The FIRMS dataset contained all the fire events from 2015-2023 with highly timed geo-specific coordinates. However, the weather data obtained from Meteostat was daily data for only the major stations in Australia. The NDVI data provided 16-day composites over the period of 2015-2023. This resulted in substantial missing values in the dataset after integration, along with gaps due to incomplete station recordings, timestamps or coordinates. Any missing values were dropped across the dataset to maintain data integrity. Features such as brightness were dropped as it was not useful for modeling. The target variable Fire Radiative Power (FRP) in the FIRMS dataset exhibited some extreme outliers probably because of human intervention. Such values were prevented by capping the FRP at 99th percentile.

\subsection{Feature Engineering}
The features were normalized and scaled wherever necessary to ensure comparability across the heterogeneous variables such as temperature, NDVI and wind speed. The feature engineering included the addition of the diurnal range which was the difference between max temperature and min temperature.
Three interaction terms were added to capture the relationships between the features.
Transformations such as sine/cosine were added to the months to record the cyclic/seasonal pattern of the months \cite{b11}.
After capping the FRP at 99th percentile, the continuous values were converted into the categorical fire risk levels. A two-class (low and high) classification was constructed, where low risk was defined as FRP$<$=40 and high risk as $>$40.
The resulting feature set combined temporal, vegetation and climatic data ensuring that the models capture both short and long-term conditions and a broader ecological context. The final integrated data was then split into test and train set with 80/20 stratification on the target risk class.
\subsection{Exploratory data analysis}
\begin{figure}
    \centering
    \includegraphics[width=1\linewidth]{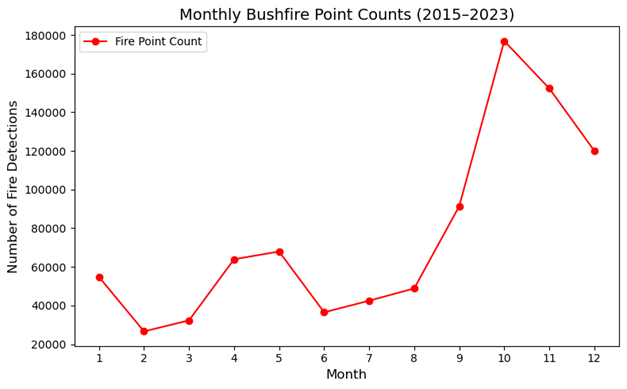}
    \caption{Monthly Bushfire Point Counts (2015–2023).}
    \label{fig:Figure 1}
\end{figure}

The temporal distribution of the bushfire counts is shown in Figure \ref{fig:Figure 1}. The trend asserts a strong seasonal pattern in the data. The X axis shows the months of the year, and the Y axis shows the frequency of fire detections. The line graph depicts the trend of the fire count across the plot.

\begin{figure}[!t]
    \centering
    \includegraphics[width=1\linewidth]{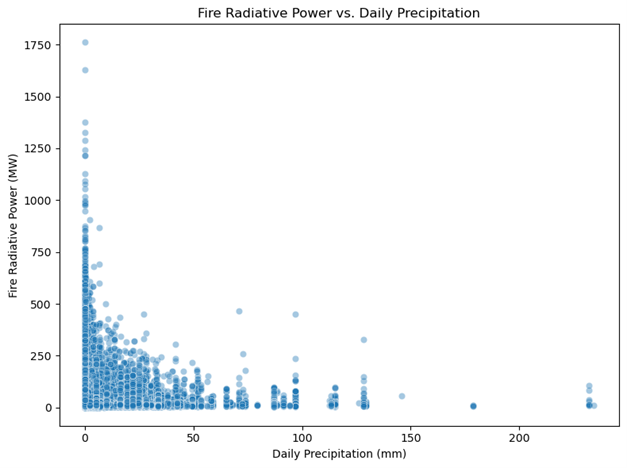}
    \caption{Fire Radiative Power (FRP) vs Daily Precipitation plot
    (2015-2023).}
    \label{fig:Figure 2}
\end{figure}

The plot of FRP vs Daily Precipitation is shown in Figure \ref{fig:Figure 2}. The x-axis contains the amount of daily precipitation, and the y-axis contains the FRP.
The plot shows a negative relationship between the two features indicating that more precipitation means less fire-prone environment. We can see a few outliers even in the high precipitation times but that might be due to human-interference or other industrial anomalies. Then again, it is very low and negligible.
During low precipitation times, we can see that there is a huge increase in FRP suggesting drier and more fire-prone conditions.

\subsection{Model Training}
Multiple machine learning models were trained to
evaluate different approaches to capture the best results for the spatio-temporal dynamics of the bushfire risk model. The machine learning models were chosen over neural networks due to better suitability for structured tabular data. The class imbalance in the datasets was not suitable for a neural network model. At first, a three-class classification was considered on all models. However, after evaluation, two-class classification was adopted to improve the generalizability of the model. Both linear and non-linear learners were considered to ensure a fair comparison of methods to achieve various levels of flexibility and interpretability.

\section{Experimental Evaluation}\label{sec:experiments}
This section presents the experimental setup used to evaluate the proposed models, dataset configurations, tools, and evaluation metrics. 

The integrated dataset including the values from 2015-2023 recorded approximately 250,000 samples. For training and testing, an 80/20 split was included to ensure proportion across the dataset. Various machine learning models were employed with a combination of hyperparameter tuning grids and class balance techniques. All experiments were conducted using standard Python libraries. The best performing model for two-class classification was recorded by performing various evaluation metrics to ensure better operational performance.

\subsection{Three-class classification results}   
Machine learning models like XGBoost, LightGBM, MLP, and Random Forest were tested to ensure a fair comparison across all
models.

\paragraph{XGBoost}
XGBoost has a reputation to work well with non-linear
interactions and captures those interactions very well. It
contains integrated early stopping and regularisations to
avoid overfitting of data \cite{b13}. XGBoost was constructed by
balancing the classes using the `RandomizedSearchCV.`
technique with a 3-fold cross-validation \cite{b14}. This resulted in a recall of 56\% for “high” class risk in the three-class classification model. It also introduced balance between classes and macro f1-score of 0.47 was obtained. The model maintained generalisation, avoiding overfitting. 

\paragraph{Random Forest classifier}
Random Forest is a decision tree technique which uses
bagging \cite{b15}. It is well suited for structured datasets like the
one used in this project. It can handle high dimensional
categorical encodings and non-linear feature interactions.
The Random Forest classifier was built using the
SMOTETomek technique which creates synthetic samples
for the minority class reducing overlaps via Tomek links \cite{b16}.
This technique mitigates class imbalance better than class
weights alone. The high-class f1-score was 24\% and the
recall was 33\%. The Macro F1-score was 47\% suggesting a
balanced performance in the model Table~\ref{tab:rf_classification_report}.
\begin{table}[htbp]
\centering
\caption{Classification Report for Random Forest Classifier (RF + SMOTETomek + Tuning)}
\label{tab:rf_classification_report}
\begin{tabular}{|l|c|c|c|c|}
\hline
\textbf{Class} & \textbf{Precision} & \textbf{Recall} & \textbf{F1-score} & \textbf{Support} \\ \hline
High   & 0.19 & 0.33 & 0.24 & 2759  \\ \hline
Low    & 0.83 & 0.79 & 0.81 & 39208 \\ \hline
Medium & 0.35 & 0.34 & 0.35 & 10149 \\ \hline
\textbf{Accuracy}     & --   & --   & \textbf{0.68} & \textbf{52116} \\ \hline
\textbf{Macro Avg}    & 0.46 & 0.49 & 0.47 & 52116 \\ \hline
\textbf{Weighted Avg} & 0.70 & 0.68 & 0.69 & 52116 \\ \hline
\end{tabular}
\end{table}

\paragraph{Multilayer Perceptron (MLP)}
The MLP is a basic neural network that accepts scaled
numeric inputs, which help model complex non-linear
patterns between the features \cite{b17}. However, it is not as
interpretable as the tree-based models. It is convenient for
comparisons in deep learning models and performs best when
combined with class-balancing techniques, as it does not
handle class imbalances well. It was trained by adding SMOTETomek analysis. Hyperparameter tuning involved the refinement of hidden layers (100, 100,50), various activation functions (relu, tanh) and different regularizations. The accuracy was around 63\% and the macro average precision and recall were around 43\% and 48\% respectively Table~\ref{tab:mlp_classification_report}.
\begin{table}[htbp]
\centering
\caption{Classification report for MLP classifier (MLP + SMOTETomek + tuning)}
\label{tab:mlp_classification_report}
\resizebox{\linewidth}{!}{
\begin{tabular}{|l|c|c|c|c|}
\hline
\textbf{Class} & \textbf{Precision} & \textbf{Recall} & \textbf{F1-score} & \textbf{Support} \\
\hline
High    & 0.16 & 0.46 & 0.23 & 2759  \\
Low     & 0.82 & 0.74 & 0.78 & 39208 \\
Medium  & 0.29 & 0.25 & 0.27 & 10149 \\
\hline
\textbf{Accuracy}     & --   & --   & 0.63 & 52116 \\
\textbf{Macro Avg}    & 0.42 & 0.48 & 0.43 & 52116 \\
\textbf{Weighted Avg} & 0.68 & 0.63 & 0.65 & 52116 \\
\hline
\end{tabular}
}
\end{table}

\paragraph{LightGBM}
The LightGBM model was tuned with `RandomizedSearchCV` with a hyperparameter tuning grid of 20 iterations, 3 learning rates (0.01, 0.05, 0.1) and various estimators (100, 200). 3-fold cross-validation was used to preserve class proportions \cite{b18}. The model performed better with balanced trade-offs between recall and precision. The tuning helped the “high” class to perform better as it is the minority class. It gave a total accuracy of 61\% and macro average F1-score of 44\% and recall of 53\% (Table~\ref{tab:lightgbm_report}).

\begin{table}[htbp]
\caption{LightGBM Classification Report}
\label{tab:lightgbm_report}
\centering
\resizebox{\linewidth}{!}{
\begin{tabular}{|c|c|c|c|c|}
\hline
\textbf{Class} & \textbf{Precision} & \textbf{Recall} & \textbf{F1-score} & \textbf{Support} \\
\hline
High    & 0.16 & 0.58 & 0.25 & 2759 \\
Low     & 0.85 & 0.69 & 0.76 & 39208 \\
Medium  & 0.31 & 0.32 & 0.31 & 10149 \\
\hline
\textbf{Accuracy}   & --   & --   & 0.61 & 52116 \\
\textbf{Macro Avg}  & 0.44 & 0.53 & 0.44 & 52116 \\
\textbf{Weighted Avg} & 0.71 & 0.61 & 0.65 & 52116 \\
\hline
\end{tabular}
}
\end{table}
\paragraph{3-class ensemble (stacked) model}
A stacked model with LightGBM and RandomForest as base learners and Linear Regression as a meta learner was used with SMOTETomek technique. The evaluation metrics improved significantly. This hybrid technique oversampled the minority class which helped in class distinction. The macro average f1-score was around 47\% which was a great improvement, and the recall was around 49\%. The accuracy was around 68\% which is not ideal, but the classes were balanced as shown in the Table~\ref{tab:ensemble3_classification_report}
.

\begin{table}[htbp]
\centering
\caption{Classification Report for Three-Class Ensemble Model (Stacked Model + SMOTETomek)}
\label{tab:ensemble3_classification_report}
\resizebox{\linewidth}{!}{
\begin{tabular}{|l|c|c|c|c|}
\hline
\textbf{Class} & \textbf{Precision} & \textbf{Recall} & \textbf{F1-score} & \textbf{Support} \\ \hline
High    & 0.19 & 0.30 & 0.23 & 2759  \\ 
Low     & 0.83 & 0.79 & 0.81 & 39208 \\ 
Medium  & 0.35 & 0.37 & 0.36 & 10149 \\ 
\hline
\textbf{Accuracy}     & --   & --   & 0.68 & 52116 \\ 
\textbf{Macro Avg}    & 0.46 & 0.49 & 0.47 & 52116 \\ 
\textbf{Weighted Avg} & 0.71 & 0.68 & 0.69 & 52116 \\ 
\hline
\end{tabular}}
\end{table}

\begin{tcolorbox}[boxrule=0.75pt]
Three-class models showed limited separability, with accuracy saturating around 68--70\%, confirming that the classes are not sufficiently distinct in real bushfire data.
\end{tcolorbox}

A complete comparison of the confusion matrices for all models, including both three-class and two-class settings, is presented in Table~\ref{tab:combined_confusion}. This consolidated table highlights the differences in misclassification patterns across the models.

\subsection{Two-class classification model}
\paragraph{2-class ensemble (stacked) model}
Finally, due to underperformance of most models on three class classifications, the problem formulation was shifted from classifying “low”, “medium” and “high” (3-class classification) to “low” and “high” (2-class classification). The two-class classification was then employed with random forest and lightGBM as baseline models with logistic regression as a
meta learner. The final accuracy was around 87\%, macro f1-score was 61\% and the recall was around 69\%. The ROC AUC was around 0.77. The detailed performance metrics of the final two-class ensemble model are summarised in Table~\ref{tab:two_class_ensemble_report}.

\begin{table}[htbp]
\centering
\caption{Two-Class Classification Ensemble Model Report (Optimized Threshold = 0.47)}
\label{tab:two_class_ensemble_report}
\resizebox{\linewidth}{!}{
\begin{tabular}{|l|c|c|c|c|}
\hline
\textbf{Class} & \textbf{Precision} & \textbf{Recall} & \textbf{F1-score} & \textbf{Support} \\ \hline
Low/Medium & 0.97 & 0.90 & 0.93 & 49357 \\ 
High       & 0.20 & 0.47 & 0.29 & 2759  \\ 
\hline
\textbf{Accuracy}     & --   & --   & 0.87 & 52116 \\ 
\textbf{Macro Avg}    & 0.59 & 0.69 & 0.61 & 52116 \\ 
\textbf{Weighted Avg} & 0.93 & 0.87 & 0.90 & 52116 \\ 
\hline
\end{tabular}}
\end{table}

\begin{table*}[!t]
\centering
\caption{Normalised Confusion Matrices for All Models Under Three-Class and Two-Class Classification (Row-wise Percentages)}
\label{tab:combined_confusion}
\resizebox{\textwidth}{!}{
\begin{tabular}{|l|c|c|c|c|c|c|c|c|c|}
\hline
\textbf{Model} & 
\textbf{H$\rightarrow$H} & \textbf{H$\rightarrow$L} & \textbf{H$\rightarrow$M} &
\textbf{L$\rightarrow$H} & \textbf{L$\rightarrow$L} & \textbf{L$\rightarrow$M} &
\textbf{M$\rightarrow$H} & \textbf{M$\rightarrow$L} & \textbf{M$\rightarrow$M} \\ \hline

XGBoost (3-Class) &
54.77\% & 21.71\% & 23.52\% &
11.02\% & 71.57\% & 17.40\% &
23.15\% & 41.10\% & 35.76\% \\ \hline

Random Forest (3-Class) &
33.06\% & 37.88\% & 29.07\% &
6.35\% & 79.16\% & 14.49\% &
14.14\% & 51.62\% & 34.24\% \\ \hline
MLP (3-Class) &
46.18\% & 34.26\% & 19.58\% &
11.76\% & 74.07\% & 14.19\% &
22.61\% & 52.69\% & 24.74\% \\ \hline

LightGBM (3-Class) &
57.63\% & 20.69\% & 21.68\% &
14.56\% & 68.51\% & 16.90\% &
27.28\% & 40.97\% & 31.77\% \\ \hline

Ensemble (3-Class) &
30.04\% & 36.12\% & 33.82\% &
5.82\% & 78.63\% & 15.59\% &
12.85\% & 50.27\% & 36.90\% \\ \hline

\textbf{Ensemble (2-Class: L/M vs H)} &
\multicolumn{6}{c|}{--- (Not applicable for 2-class)} &
52.71\% (H) & 10.25\% (H) & 89.75\% (L/M) \\ \hline

\end{tabular}}
\end{table*}

\begin{figure}[!t]
    \centering
    \includegraphics[width=1\linewidth]{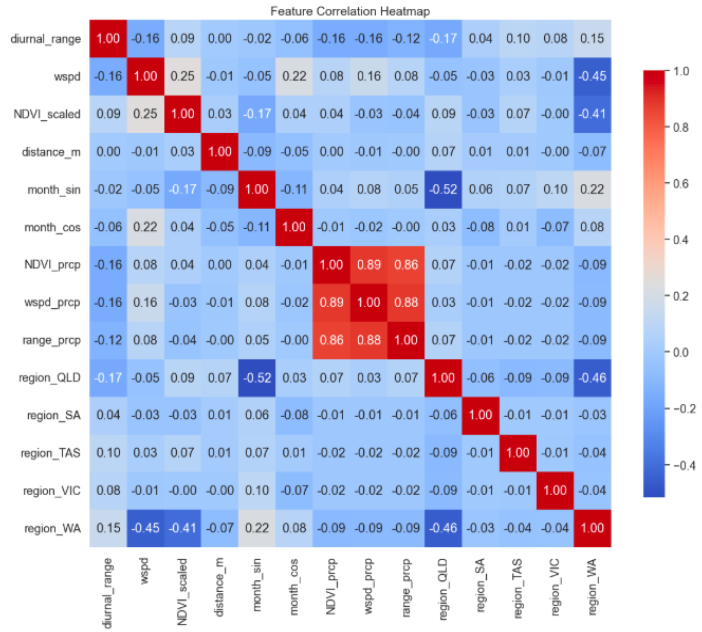}
    \caption{Correlation Heatmap of the Two-class Classification Technique with Ensemble Model.}
    \label{fig: Figure 16}
\end{figure}

The final two-class classification model used a Stacking Classifier combining random forest, lightGBM and linear regression. This stack was embedded with SMOTE pipeline to ensure class balance in the model. Unlike default conventional pipelines which use 0.5 as threshold for classification, this model used 0.47 maximizing the f1-score for high-risk fire predictions. The consequences of misidentified data are too severe, hence the adjustment. At final threshold, the model gave 87.49\% accuracy with 47.3\% recall and 28.59\% f1 score with a precision of 20.49\%. The recall is more prioritized in environmental data as false positives are better than misclassifications. False positives can mean deployment of countermeasures which might not be required but false negatives could cause a catastrophic event without any warning \cite{b20}. The correlation heatmap ensures that while interaction terms have internal correlations, the primary features and dummy regions remain uncorrelated \cite{b21}. This ensures robust feature learning, as shown in Figure \ref{fig: Figure 16}

\begin{tcolorbox}[boxrule=0.75pt]
Across all models, high-risk fires remain the hardest to classify, reinforcing the need for recall-focused evaluation in bushfire prediction.
\end{tcolorbox}

\section{Discussion} \label{sec:discussion}
This section highlights experimental results and their broader implications for both practitioners and researchers working in bushfire prediction. 

In the initial stage of this project, the task was formulated as three-class classification. However, the accuracy saturated at 68-70\% across all models and hyperparameter configurations. This behaviour is common in real-world environmental data because classes are poorly separated. In this project, a fire occurrence is very sparse (1 in 5000), which causes class imbalances. There may be instances where a fire is triggered by a lightning strike in high temperature environment with dry vegetation but most other times, without the lightning in the region, even high temperatures and dry vegetation might not spike fires. Constructing a model with such parameters is very challenging. This makes three-class classification difficult. In general, a VIF check confirms, and it is preferred to be less than 5 but, in environmental data the level of multicollinearity achieved in this project is not harmful \cite{b22}. We can also observe that the multicollinearity again is only by the interaction terms.

\begin{tcolorbox}[boxrule=0.75pt]
Simplifying the problem to two classes produced more stable predictions and improved practical reliability for bushfire early-warning systems.
\end{tcolorbox}

For researchers, this study provides a reproducible FIRMS-Meteostat-NDVI framework that can serve as a baseline for bushfire modeling, incorporating additional indices, fuel-related variables, and advanced spatio-temporal models. The two-class ensemble provides a simple, open data-driven tool to flag high-intensity fire risk and support operational planning. From a research perspective, this baseline pipeline shows how openly available fire, meteorological, and vegetation data can be integrated into a single, reproducible workflow. 
The two-class ensemble provides a simple, open data-driven tool to flag high-intensity fire risk and support operational planning. In practice, it can help agencies prioritise high-risk areas and support early warning and preparedness decisions using routinely available data. Table ~\ref{tab:vif_values} reports the Variance inflation factors (VIF) for the final two-class classification model, indicating the multicollinearity is largely confined to interaction terms. 

\begin{table}[htbp]
\caption{Variance Inflation Factors (VIF) of the Final Two-Class Classification Model}
\label{tab:vif_values}
\centering
\resizebox{\linewidth}{!}{
\begin{tabular}{|l|c|l|c|}
\hline
\textbf{Feature} & \textbf{VIF} & \textbf{Feature} & \textbf{VIF} \\
\hline
wspd\_prcp      & 6.887279  & month\_sin      & 1.473759 \\
NDVI\_prcp      & 5.612261  & NDVI\_scaled    & 1.349871 \\
range\_prcp     & 5.382169  & month\_cos      & 1.172513 \\
region\_WA      & 2.286368  & diurnal\_range  & 1.142786 \\
region\_QLD     & 2.036589  & region\_VIC     & 1.034151 \\
wspd            & 1.732007  & region\_TAS     & 1.033466 \\
region\_SA      & 1.023343  & distance\_m     & 1.016756 \\
\hline
\end{tabular}
}
\end{table}

\begin{figure}[!t]
    \centering
    \includegraphics[width=1\linewidth]{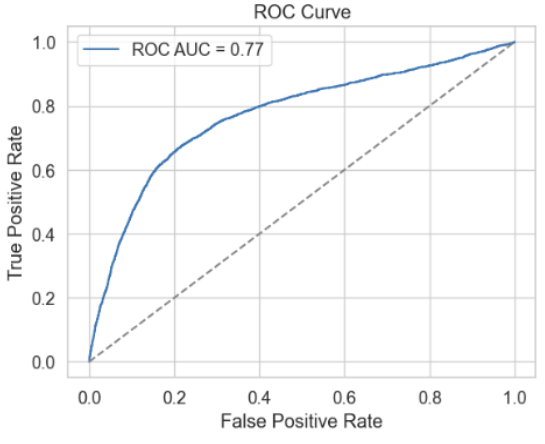}
    \caption{ROC-AUC Curve for the Final Two-class Ensemble Model.}
    \label{fig: Figure 18}
\end{figure}

A binary classification was found to be a more grounded approach for this project. By merging low/medium classes into one, the noise is reduced significantly, and there is more distinction. The simplification was also justified by ROC-AUC (0.77) which is a very critical metric for emergency response models (Figure \ref{fig: Figure 18}).

This study demonstrates the possibility of merging various datasets both spatially and temporally to study bushfires and predict them across Australia.
For researchers, the findings highlight several directions for future work. The use of an ensemble model shows that the limitations of the individual models can be broken to achieve better results. The moderate performance of the three-class classification indicates the missing data that is favoring the bushfires. This project only integrates the vegetation, climatic and fire data which makes it evident that other than these factors there are other missing features causing fires. Acquisition of other indices like drought, bio-fuel index, human intervention etc. can lead to promising results.
For practitioners, an effective two-class model with accurate results provides decent predictions. The emphasis on recall ensures that high risk events are less likely to be overlooked. Furthermore, the platform is built on open source datasets which can easily be replicated.

\section{Threats to Validity} \label{sec:limitations}
This section outlines the key limitations and potential threats to validity associated with the study. The feature scope is limited to NDVI, as additional vegetation indices like the NBR, NDWI, and EVI were not included in the project. This decision maintains workflow efficiency while capturing crucial vegetation behavior related to fire activity. Nonetheless, NDVI might be insufficient to capture all the information that might be complemented by other indices, and, as a result, reduce the whole vegetation-fire characterization.

Spatial and temporal joins between FIRMS, Meteostat, and NDVI resulted in the removal of unmatched records due to timing and resolution differences. Although this reduced the total sample size, it ensured higher data consistency and prevented noisy or incomplete observations from influencing the model. This filtering step may also introduce a degree of selection bias towards well-observed events and limit the representativeness of the dataset. 

Finally, computational and modeling constraints limited due to the use of more complex spatio-temporal deep learning methods, leading to the adoption of tabular machine learning models and ensembles. Despite this, the selected models provided stable performance and formed a strong baseline for future extensions. As a result, the generalisability of the workflow to other regions, sensor types, or more complex fire regimes should be interpreted with caution.

\section{Conclusion} \label{sec:conclusion}

We proposed a spatio-temporal modeling approach to integrate multi-source environmental datasets and evaluate their ability to predict bushfire risk zones across Australia.  
NASA’s FIRMS, Meteostat, and GEE’s NDVI data for the period 2015-2023 were integrated to develop a model involving spatial and temporal joins. It was evaluated on both three-class and two-class classifications via various machine learning models. The three-class classification did not generalize well and gave an accuracy of 68\%. By contrast, the two-class classification model prioritizing recall generalised well to achieve an accuracy of 87\%.
The findings show that integrating environmental datasets provides a promising and practical approach to bushfire predictions.
Future extensions can incorporate additional datasets like drought and other vegetation indices, along with human activity data, to provide better results. Overall, this project demonstrates a feasible step towards data-driven bushfire risk assessment for disaster management applications.

\section{Replication Package}
We provide the replication package \cite{b32} to reproduce the analysis and results reported in this paper.



\bibliographystyle{IEEEtran}
\bibliography{main}

\end{document}